\documentclass[conference]{IEEEtran}
\usepackage{cite}
\usepackage{amsmath,amssymb,amsfonts}
\usepackage{algorithmic}
\usepackage{graphicx}
\usepackage{textcomp}
\usepackage{xcolor}
\usepackage[colorlinks=true, urlcolor=blue, linkcolor=red]{hyperref}
\newcommand{\figref}[1]{Fig.~\ref{#1}}
\usepackage{tikz}
\usepackage{textcomp}

\renewcommand{\eqref}[1]{eq.~\ref{#1}}

\newcommand{\secref}[1]{\S \ref{#1}}

\newcommand{\rebut}[1]{\textcolor{black}{#1}}

\def\BibTeX{{\rm B\kern-.05em{\sc i\kern-.025em b}\kern-.08em
    T\kern-.1667em\lower.7ex\hbox{E}\kern-.125emX}}

\newcommand\copyrighttext{%
  \footnotesize \textcopyright 2024 IEEE.  Personal use of this material is permitted.  Permission from IEEE must be obtained for all other uses, in any current or future media, including reprinting/republishing this material for advertising or promotional purposes, creating new collective works, for resale or redistribution to servers or lists, or reuse of any copyrighted component of this work in other works.}
\newcommand\copyrightnotice{%
\begin{tikzpicture}[remember picture,overlay]
\node[anchor=south,yshift=10pt] at (current page.south) {\fbox{\parbox{\dimexpr\textwidth-\fboxsep-\fboxrule\relax}{\copyrighttext}}};
\end{tikzpicture}%
}

\begin{document}

\title{Learning Object Semantic Similarity with Self-Supervision}


\author{
    \IEEEauthorblockN{
        Arthur Aubret$^{1,2,}$\IEEEauthorrefmark{1}, Timothy Schaumlöffel$^{3,4,}$\IEEEauthorrefmark{1}, Gemma Roig$^{3,4,}$\IEEEauthorrefmark{2}, Jochen Triesch$^{1,3,}$\IEEEauthorrefmark{2}
    }
    \IEEEauthorblockA{$^1$Frankfurt Institute for Advanced Studies, Frankfurt a. M.}
    \IEEEauthorblockA{$^2$Xidian-FIAS international Joint Research Center, Frankfurt a. M.}
    \IEEEauthorblockA{$^3$Goethe University, Frankfurt a. M.}
    \IEEEauthorblockA{$^4$The Hessian Center for Artificial Intelligence, Darmstadt}
    \thanks{\IEEEauthorrefmark{1}Shared first authorship, \IEEEauthorrefmark{2}shared last authorship.}
}


\maketitle
\copyrightnotice

\begin{abstract}
Humans judge the similarity of two objects not just based on their visual appearance but also based on their semantic relatedness. However, it remains unclear how humans learn about semantic relationships between objects and categories. One important source of semantic knowledge is that semantically related objects frequently co-occur in the same context. For instance, forks and plates are perceived as similar, at least in part, because they are often experienced together in a ``kitchen" or ``eating'' context.  Here, we investigate whether a bio-inspired learning principle exploiting such co-occurrence statistics suffices to learn a semantically structured object representation {\em de novo} from raw visual or combined visual and linguistic input. To this end, we simulate temporal sequences of visual experience by binding together short video clips of real-world scenes showing objects in different contexts. A bio-inspired neural network model aligns close-in-time visual representations while also aligning visual and category label representations to simulate visuo-language alignment. Our results show that our model clusters object representations based on their context, e.g.\ kitchen or bedroom, in particular in high-level layers of the network, akin to humans. In contrast, lower-level layers tend to better reflect object identity or category. To achieve this, the model exploits two distinct strategies: the visuo-language alignment ensures that different objects of the same category are represented similarly, whereas the temporal alignment leverages that objects from the same context are frequently seen in succession to make their representations more similar. Overall, our work suggests temporal and visuo-language alignment as plausible computational principles for explaining the origins of certain forms of semantic knowledge in humans.
\end{abstract}



\section{Introduction}

Humans learn visual representations that allow them to judge the similarity between different images of objects. These representations typically rely on learnt semantic attributes, like the identity of an object or its category or the semantic relationships among different categories. A recent study indicates that the visual context in which an object commonly appears may also shape its perceived similarity with other objects, even when evaluated in the absence of background \cite{turini2022hierarchical}. This is in line with research showing that an image of an object, even without backgrounds, can elicit activity in brain areas dedicated to scene representations \cite{troiani2014multiple,harel2013deconstructing}.
For instance, the concepts ``mug''  and ``coffee maker'' may be judged similar because specific instances naturally and frequently co-locate in a ``kitchen" context. However, the learning principles underpinning the formation of such semantically structured representations remain unclear.
Biological systems continually interact with the world and objects therein, which elicits a spatio-temporal structure in their visual experience. On the object level, our visual experience entails two kinds of statistical co-occurrences: 1) spatial ones refer to co-occurrences of an object with elements of the background and 2) ``temporal" ones indicate that a specific object will likely be seen after another specific one. Contexts clearly define spatio-temporal co-occurrences between objects. Imagine a toddler in its home. During lunchtime, he eats in the kitchen and alternately regards his mug, a coffee maker, and the food on the table, driving many visual transitions between these objects. Then, he is taken to the bathroom to wash his hands and successively observes a sink, tap, and soap. This example of a natural succession of events entails two distinct temporal clusters of visual transitions, \textit{i.e.} those occurring in the ``kitchen" context and those in the ``bathroom" context. 

A long-standing hypothesis states that the brain learns to capture redundancies in visual statistics progressively \cite{simoncelli2001natural,barlow1961possible}. Thus, biological systems may actually leverage the structure of their visual experience to build more efficient visual representations \cite{turk2019hippocampus,miyashita1988neuronal,schapiro2012shaping,schapiro2016statistical}. For instance, the slowness principle states that biological systems exploit the temporal structure of their visual experience in order to map close-in-time visual inputs to similar representations \cite{wiskott2002slow}.

\begin{figure*}[ht]
    \centering
    \includegraphics[width=1\linewidth]{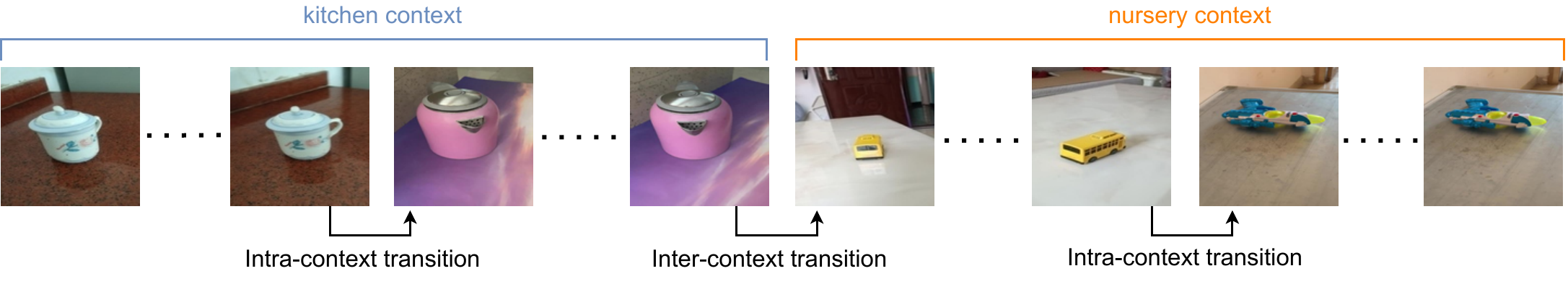}
    \caption{We simulate extended egocentric visual experience of objects in different contexts. Objects from the same context have a high probability of being seen in succession (intra-context transition), while this probability is reduced for objects from different contexts (inter-context transition).}
    \label{fig:temporal_sequence}
\end{figure*}
\begin{figure}
    \centering
    \includegraphics[width=1\linewidth]{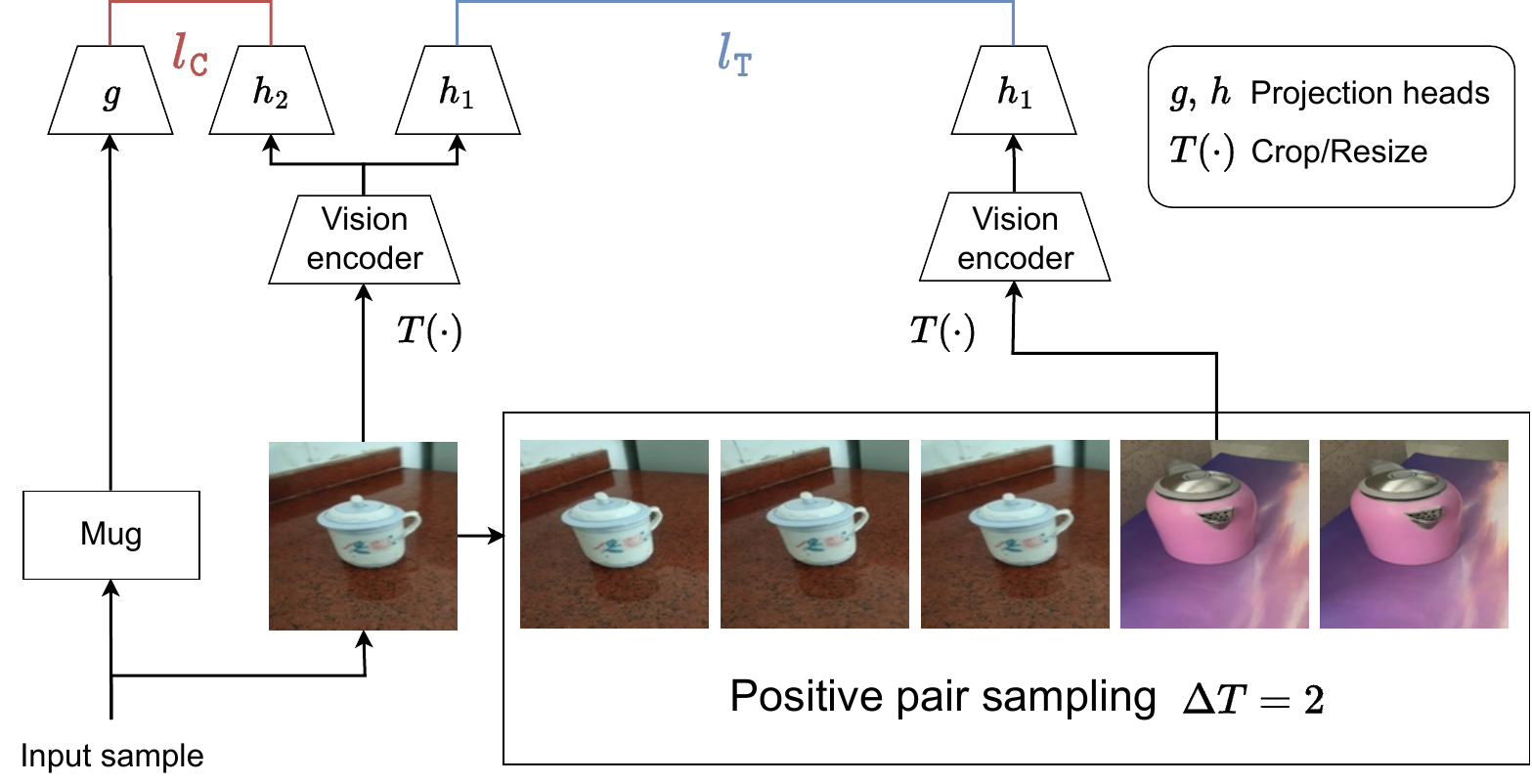}
    \caption{Learning architecture. See text for details.}
    \label{fig:learnarch}
\end{figure}

In this paper, we investigate whether the slowness principle and a visuo-language alignment suffice for establishing a context-wise organization of object representations from an egocentric experience of the world. We simulate sequences of egocentric visual experience using single-object images extracted from a large-scale real-world image dataset \cite{yu2023mvimgnet}. The sequences present temporally consistent images of object instances linked by between-object transitions. We distinguish (more frequent) {\em intra-context transitions} between, e.g., mug and kettle from the same ``kitchen'' context from (less frequent) {\em inter-context transitions}, e.g., from mug (``kitchen'' context) to toys (``nursery context''). Then, we use a bio-inspired learning model that 1) \rebut{implements the slowness principle by mapping} close-in-time visual inputs to similar representations and 2) approximates visuo-language alignment learning by making similar image representations and representations of their respective object category labels. Our data sequence and learning architecture are displayed in \figref{fig:temporal_sequence} and \figref{fig:learnarch}.

Our results show that high-level layers of the network cluster the visual representations in a context-wise fashion, in particular when the visual sequence shows few inter-context transitions, similar to biological systems. Our analysis emphasizes that the level of sparsity increases through different layers, driving a trade-off between category-wise and context-wise structure in the representations.
An ablation study further reveals that the temporal slowness objective captures both spatial and temporal co-occurrences. In contrast, our approximated language supervision only extracts spatial co-occurrences from an image. Overall, our work sheds light on the potential role of two learning principles, temporal slowness and multi-modal alignment, in constructing semantic mental representations in humans. Code and data will be available at \href{https://github.com/neuroai-arena/ObjectSemanticSimilarity}{https://github.com/neuroai-arena/ObjectSemanticSimilarity}.










\section{Related work}

\paragraph{Slowly changing visual representations}

Previous computational models implementing the slowness principle learnt visual representations suitable for recognizing object instances \cite{schneider2021contrastive,franzius2011invariant}, categorizing objects \cite{parthasarathy2023self,aubrettime,orhan2020self}, recognizing high-level time-extended actions \cite{feichtenhofer2021large,knights2021temporally} and understanding scenes \cite{parthasarathy2023self,gordon2020watching}. When learning on image sequences showing natural interactions with objects, these models mostly learn viewpoint/position invariant object representations \cite{schneider2021contrastive,yu2023mvimgnet}. Furthermore, thanks to their bio-plausibility, slowness objectives are increasingly used to model the development of visual representations in infants \cite{schaumloffel2023caregiver,aubret2022toddler,sheybani2023curriculum}. To the best of our knowledge, we are the first to rigorously study the impact of context transitions on representations learnt through the slowness principle. 

\paragraph{Impact of visual co-occurrences on object representations}

To learn object embeddings accounting for visual co-occurrences, \cite{bonner2021object} consider a scene dataset where all objects are labeled and learn embeddings by predicting objects present in a scene of a given object concept, in a word2vec fashion \cite{mikolov2013distributed}. The learnt embeddings can predict neural activity in the parahippocampal place area when subjects are seeing individual objects, an area commonly associated with scene recognition. In contrast to this approach, we learn representations directly from pixels and rigorously study how context-wise visual representations emerge. In addition, it remains unclear how the spatial co-occurences studied in \cite{bonner2021object} align with natural temporal co-occurrences. Other works compare representations learnt by a convolutional neural network (CNN) trained in a supervised fashion with human representations. Among them, one showed that contextual object prototypes, built by averaging scene representations containing a given object, capture judgements of visual similarity \cite{magri2022context}. Furthermore, other work highlights that a CNN embedding groups together objects that occur in similar contexts \cite{turini2022hierarchical} as well as objects and their typical context \cite{bracci2021representational}. Rather, we focus on how a bio-inspired learning mechanism based on temporal slowness can induce a context-wise organization of visual representations. \rebut{A distinct line of works also studies object relation modeling within a scene}\cite{bozcan2019cosmo}.

\section{Methods}\label{sec:methods}

We investigate whether a bio-inspired model endowed with slowness constraints can learn context-based semantic representations from a natural sequence of egocentric visual inputs. We first explain how we build a temporal sequence of images in \secref{sec:dataset} and then describe our learning model in \secref{sec:ssl}.

\subsection{Temporal Sequences of Egocentric Visual Inputs}\label{sec:dataset}

\paragraph{Dataset} \rebut{To mimic a temporal sequence of visual input experienced by a human and evaluate the learnt representation, we leverage a large dataset of video clips containing real-world objects annotated with category labels.} 
The MVImageNet dataset\cite{yu2023mvimgnet} contains 219,188 video clips that simulate a person watching and turning around a single object. Each clip contains approximately 30 frames and all objects are distributed into 238 human-centric categories. 


The MVImageNet dataset does not contain information about the context/background of an image. Thus, we define 8 distinct in-door contexts (``kitchen", ``living room", ``bedroom", ``bathroom", ``nursery", ``hallway", ``office" and ``garage") and manually allocate each object category to the context that most often contains this category of objects (as judged by the authors). We manually remove categories that do not clearly belong to one of the contexts or contain too few samples ($<100$). We further randomly remove categories until the dataset contains a balanced number of categories per context and sample a balanced number of objects per category, resulting in 80 categories and approximately 391,245 train images and 255,395 test images. We also consider a random assignment of the selected categories in \secref{sec:expe}.


\paragraph{Creation of Temporal Sequences of Images}

We create temporal sequences by progressively appending chunks of video clips. The dataset contains only one video clip per object. However, in domestic environments, humans often observe the same object several times. To account for that, we divide the original video clips into several smaller clips, by sampling section points using a Poisson distribution ($\gamma=8$); on average, this separates each original clip into four different clips. In the following, by ``clips", we refer to our crops of the original video clips.

We want to study if/how the frequency of inter-context vs.\ intra-context transitions impacts the learning process of our model. Thus, we carefully construct temporal sequences with different probabilities of a context change $p_c$. First, we randomly sample a context and a clip belonging to that context. Second, we simulate a possible context transition (e.g.\ leaving the kitchen to go to the living room) and choose with probability $p_c$ if the future clip will belong to a different context than the current one. If it belongs to the same context, we sample a new clip in the same context; otherwise, we apply a sampling weighted by the number of non-sampled clips to obtain another context and randomly sample a new clip in that context. The sampled clip is appended to the sequence and we repeat the second step until 80\% of the training clips have been added.
Without such a threshold, it may happen that all clips from all but one context have been sampled, which prevents context changes to occur and introduces a difference between the desired $p_c$ and the actual one. 

\figref{fig:temporal_sequence} shows an example of a small temporal sequence. It simulates an observer that watches a mug in a kitchen, before looking at a kettle. Then, he goes to the nursery and sees a toy bus, and finally a water pistol.

\rebut{Some experiments suggest that saccadic eye movements combined with bio-inspired cortical magnification have a similar effect to the widespread crop/resize augmentation \cite{wang2021use}.} Thus, we emulate the temporal variation induced by saccades by randomly cropping/resizing the input images during training, with the extent of cropping varying between 50\% and 100\% of the original image size.








\subsection{Representation Learning}
\label{sec:ssl}

The model learns representations by minimizing two loss functions. First, it implements learning via the slowness principle using SimCLR-TT \cite{schneider2021contrastive, aubrettime}, which allows it to map close-in-time visual inputs to similar representations. Second, we simulate learning visuo-language alignment by making similar a category representation (mimicking linguistic input provided externally) and its associated visual representations. Previous work suggests that developmentally plausible visuo-language co-occurrences combined with a learning model that aligns the two modalities can boost object categorization \cite{schaumloffel2023caregiver}. In the current work, this multimodal alignment is important to ensure that different instances of, e.g., the category ``cup'' are represented similarly. \figref{fig:learnarch} sums up our learning architecture.

\subsubsection{Self-Supervised Learning Through Time (SSLTT)} \label{sec:cltt}
At each stage of the training process, we randomly sample a mini-batch of images, denoted as $\mathcal{X}$. For each image $x_i$ we randomly select either its predecessor or successor within a time window of $\Delta t=2$ steps. The image $x_j$ selected in this way can represent the following object scenarios: 
1. the same object viewed from different orientations and positions, 2. if the clip ends, with probability $p_c$ a different object from an alternative context or, 3. with probability $1 - p_c$ a different object from the same context.
We extract features $z_i^1 = h_1(f(x_i))$ using a convolutional neural network architecture denoted as $f$ and a multi-layer perceptron (MLP) $h_1$ \cite{chen2020simple}. Finally, for a pair $(z_i^1, z_j^1)$, we compute the loss:
\begin{equation}
       l_{\texttt{T}}(z^1_i,z^1_j) = -\log \frac{\exp\left(\cos(z^1_i,z^1_j)/\tau\right)}{\sum_{\substack{z \in \mathcal{Z}^1, z \neq z_i^1}} \exp \left( \cos(z_i^1, z) / \tau \right)} \, ,
        \label{eq:simclrtt}
\end{equation}
where, $\cos$ represents the cosine similarity, $Z^1$ encompasses all embeddings $z^1$, and $\tau$ denotes the temperature hyper-parameter \cite{chen2020simple}. The numerator of \eqref{eq:simclrtt} is designed to bring the embeddings of temporally close images closer together, while the denominator of \eqref{eq:simclrtt} ensures that all image embeddings stay distinct from one another.

\subsubsection{Simulating Language Guidance} \label{sec:cat}
We use the category label $c_{i}$ for each sampled image $x_{i}$, aiming to mimic linguistic guidance. The category $c_{i}$ is represented in a one-hot scheme, spanning all 80 categories, and is processed through a neural network $g$ to produce a categorical representation $z_i^3 = g(c_1)$. 
Subsequently, we compute the embeddings for the visual representations $z_i^2 = h_2(f(x_i))$ within a unified feature space, utilizing an additional projection head, $h_2$. The objective is to minimize the distance between each corresponding pair $(z_i^2, z_i^3)$ by optimizing the following loss:
\begin{equation}
       l_{\texttt{C}}(z^2_i,z^3_i) = -\log \frac{\exp\left(\cos(z^2_i,z^3_i)/\tau\right)}{\sum_{\substack{z \in \mathcal{Z}^2, z \neq z_i^2}} \exp \left( \cos(z_i^2, z) / \tau \right)} \, ,
        \label{eq:simclrcat}
\end{equation}
where $\mathcal{Z}^2$ contains the provided embeddings $z^2$ and $z^3$.  

The total loss function $\mathcal{L}$ for a batch of size $N$ is the symmetric and batch-wise sum of \eqref{eq:simclrtt} and \eqref{eq:simclrcat}.

\subsection{Training and Evaluation}

\paragraph{Training} we use a ResNet50 \cite{he2016deep} as vision encoder $f$, and a fully connected neural network with two hidden layers of size $1024$ followed by batch normalization and ReLU activation for the category encoder $g$ and the projection heads $h_1$ and $h_2$. Each projection head projects its input representation into an embedding space of dimension $512$. We train our model for 100 epochs using the AdamW optimizer with a constant learning rate of $1e-3$, a weight decay of $1e-6$ and a batch size of 512. The temperature parameter $\tau$ is set at $0.5$ for SSLTT and $0.1$ for language guidance, based on initial experiments with $\tau \in \{0.07, 0.1, 0.5, 1\}$. \rebut{Unless stated otherwise, we select $p_c=0.1$.}
 
\paragraph{Evaluation} 
we adopt an accuracy metric that matches the odd-one-out (OOO) task used to collect similarity judgements in humans \cite{turini2022hierarchical}. For each embedding of a test image, we randomly sample one test embedding from the same context and another one from a different context and compute the cosine similarity between each pair of the triplet. We derive an accuracy metric by normalizing how often the ``same context" similarity is higher than the two others. We used two random seeds in our main experiments and observed a very small standard deviation. Thus, we only used one seed in the remaining experiments to save computational resources.

We evaluate several layers of the model. In the following, ``Representations" refers to the output of the vision encoder $f$. 
Additionally, ``VLA (\textit{n})" and ``SSLTT (\textit{n})", with $n \in \{1, 2\}$, correspond to the outputs from the first (1) or second (2) ReLU activation within the visual-language-alignment (VLA) or SSLTT projection heads, respectively, defined as $h_1$ and $h_2$.


\section{Experiments}\label{sec:expe}

Our experiments aim to assess how the spatio-temporal structure of visual experience can inject semantic knowledge into learnt representations. \rebut{First, we assess the importance of temporal object co-occurrences for learning context-wise representation. Then, we study the strategy of our model and analyze the relation between context and other semantic attributes like object identity or category. Finally, we study how temporal correlations affect the learnt representations. }

\subsection{Temporal Slowness Suffices for Capturing Contextual Structure}

\begin{figure}
    \centering
    \includegraphics[width=\linewidth]{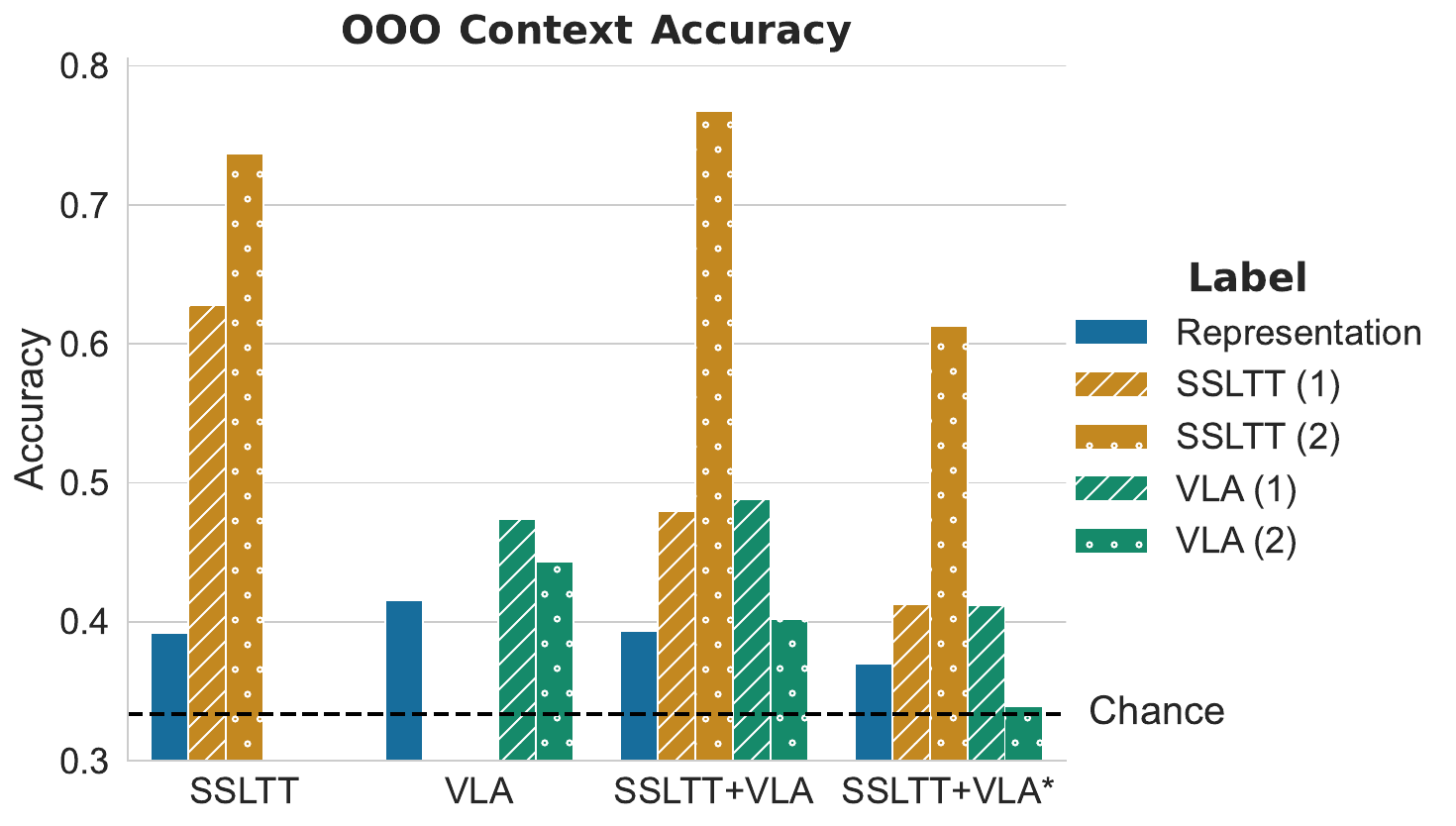}
    \caption{Differently trained models evaluated with odd-one-out test accuracy on randomly assigned context labels. The models utilize either one of the two losses or both at the same time. ``SSLTT-VLA*" denotes SSLTT-VLA trained and tested with random context assignments. The labels refer to the evaluated layer of the network.}
    \label{fig:net_type}
\end{figure}

\rebut{In \figref{fig:net_type}, we first compare our model SSLTT+VLA to a baseline that does not leverage temporal correlations (VLA). We clearly observe that the most abstract hidden layer of the projection head of SSLTT (\textit{SSLTT (2)}) learns between-object similarity that matches the frequently co-occurring context of objects. In comparison, the different layers of the comparison baseline VLA perform much worse. This highlights the importance of temporal slowness, as this layer is the closest to the layer on which we apply SSLTT.}

\begin{figure}
    \centering
    \includegraphics[width=\linewidth]{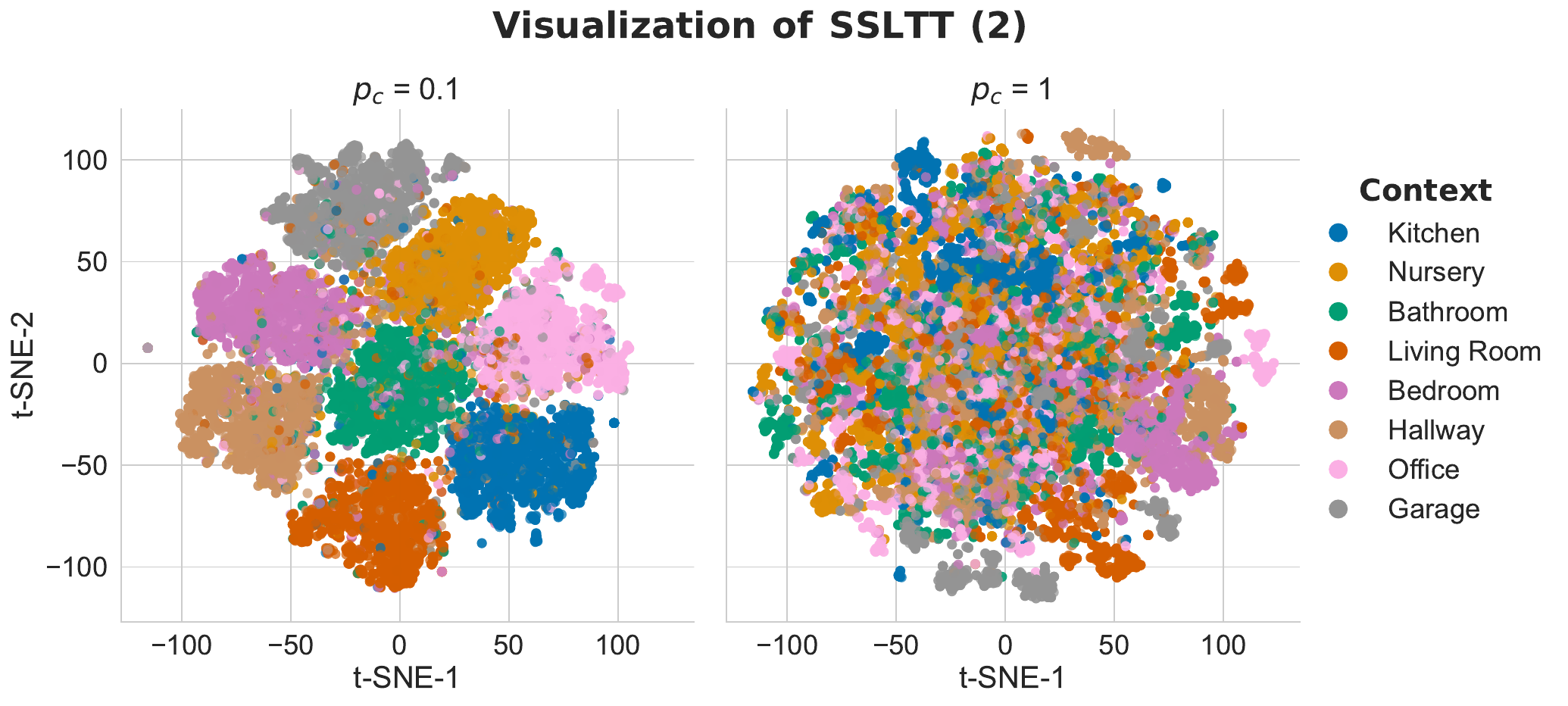}
    \caption{Visualization of the image representations using random images from the test dataset on the second layer of $h_1$. The colors represent the eight different contexts. The left plot shows the model trained using an image sequence with $p_c=0.1$, whereas the right plot uses $p_c=1.0$.
    }
    \label{fig:tsne_p}
\end{figure}

To qualitatively verify this result, we apply a t-SNE on the most abstract hidden layer of the projection head of SSLTT (\figref{fig:tsne_p}). We clearly identify 8 different clusters that correspond to the 8 contexts when inter-context transitions are infrequent ($p_c = 0.1$) but not when they are frequent ($p_c = 1.0$). We conclude that the model leverages the spatio-temporal structure of its visual experience to learn object representations that are organized by the context in which objects typically occur.

\subsection{\rebut{Our model builds context-wise representations without relying on background cues}}

Our model may leverage two different strategies to capture object co-occurrences. It may either detect that two objects often share the same backgrounds or leverage the temporal correlations. Here, we aim to disentangle the relative importance of each strategy. To prevent the model from using the background, we replace our category-to-context assignment with a random assignment (SSLTT+VLA*) \rebut{during training and testing}. \figref{fig:net_type} shows that the highest-level layer of the visuo-language part of the model VLA(2) can no longer build context-wise representations. This differs from the projection heads of SSLTT, which manage to build structured representation on par with the method trained on the original assignments (SSLTT+VLA). However, we observe a slight decrease in accuracy, suggesting that it was also benefiting from object-background co-occurrences. We deduce that, unlike visuo-language alignment, temporal slowness constructs context-wise representations using the two strategies.




\subsection{Different Network Layers Encode Different Kinds of Semantic Information}

\begin{figure}
    \centering
    \includegraphics[width=\linewidth]{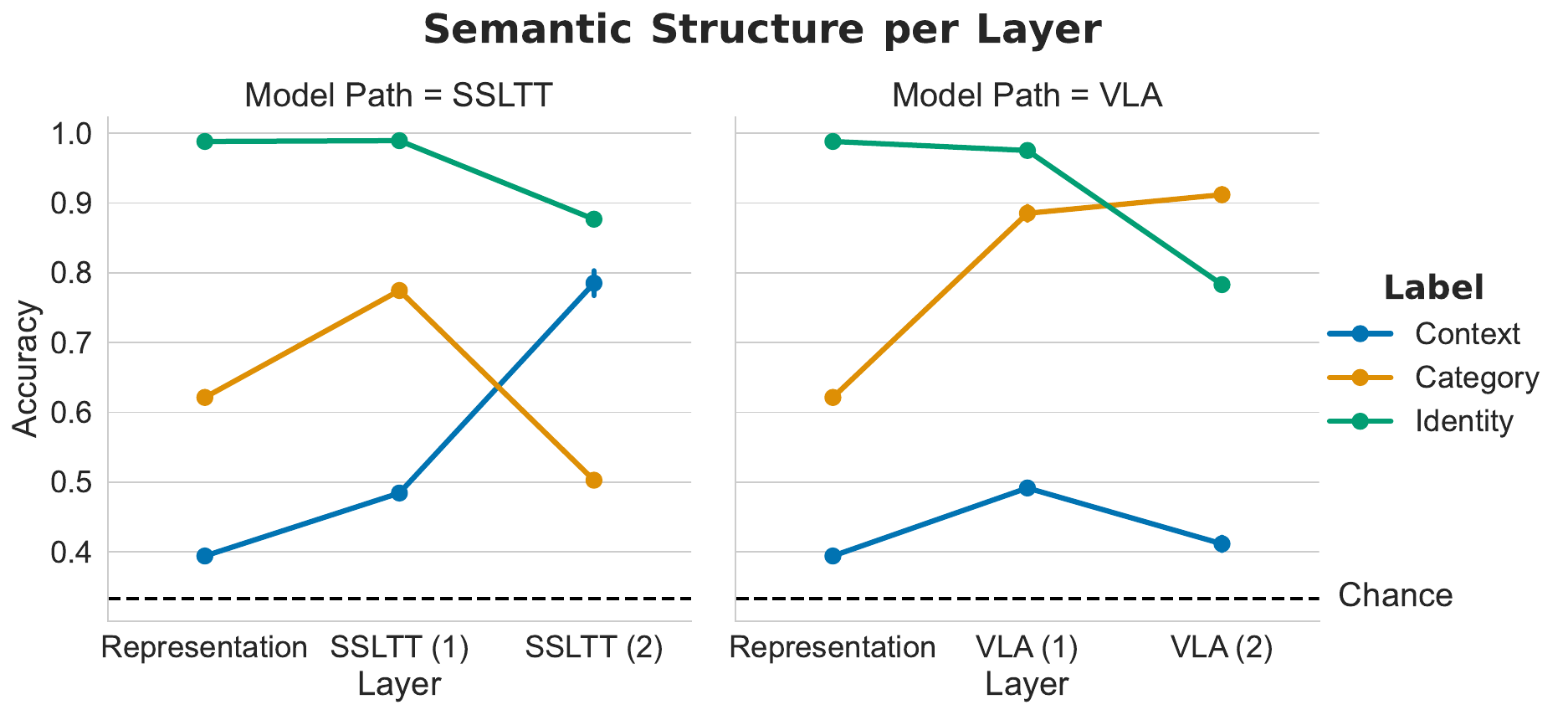}
    \caption{Odd-one-out test accuracy for context, category and object instance labels measured at different levels of the network.}
    \label{fig:net_depth}
\end{figure}

\begin{figure}
    \centering
    \includegraphics[width=1\linewidth]{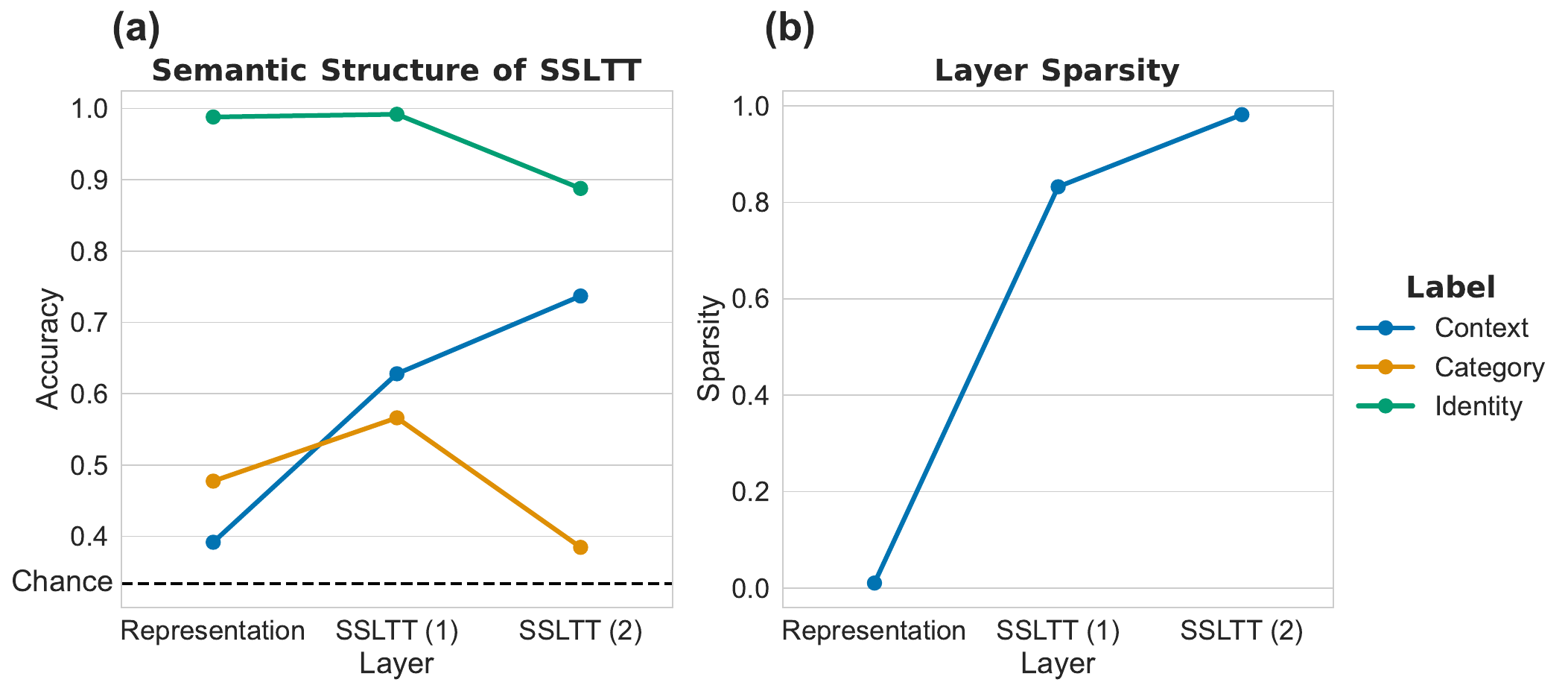}
    \caption{(a) Odd-one-out test accuracy for different label types of a model trained only on SSLTT. (b) Sparsity of the network, measured as the average percentage of neurons with zero activity per layer.}
    \label{fig:layer_sparsity}
\end{figure}

Here, we study how the model trades off representing different semantic attributes, namely an object's identity, its category and its context in the different layers of the network. In \figref{fig:net_depth}, we see that the different semantic attributes are preferentially represented in different layers. While the output of the vision encoder is ideally suited for retrieving object identity, the two layers of the SSLTT projection head are best for retrieving object category and object context, respectively. Interestingly, we do not observe this phenomenon in the visuo-language projection head. Yet, we still observe a trade-off between, on one side, category and, on the other side, object identity and context. In \figref{fig:layer_sparsity}a), we replicate the previous study with a model learnt without visuo-language guidance. We observe a similar trend as for \figref{fig:net_depth}, thereby invalidating the hypothesis that this trade-off stems from using two different loss functions. 

To further investigate these results, we assess the sparsity of the layers in \figref{fig:layer_sparsity}b). We clearly see that the highest-level layer is largely sparser than the previous one, despite those two having the same number of neurons. This suggests that, along SSLTT, an increasing compression of the visual input favors the extraction of an object context. Previous studies already noted the progressive compression effect of self-supervised learning \cite{wen2021toward,ben2023reverse}, but its origin remains unclear. Overall, we conclude that the model learns a natural trade-off between category and context structures, which may be driven by an increased sparsity across the network.


\subsection{Abundance of Context Switches Harms Context-wise Organization of Representations}

\begin{figure}
    \centering
     \includegraphics[width=\linewidth,height=6cm]{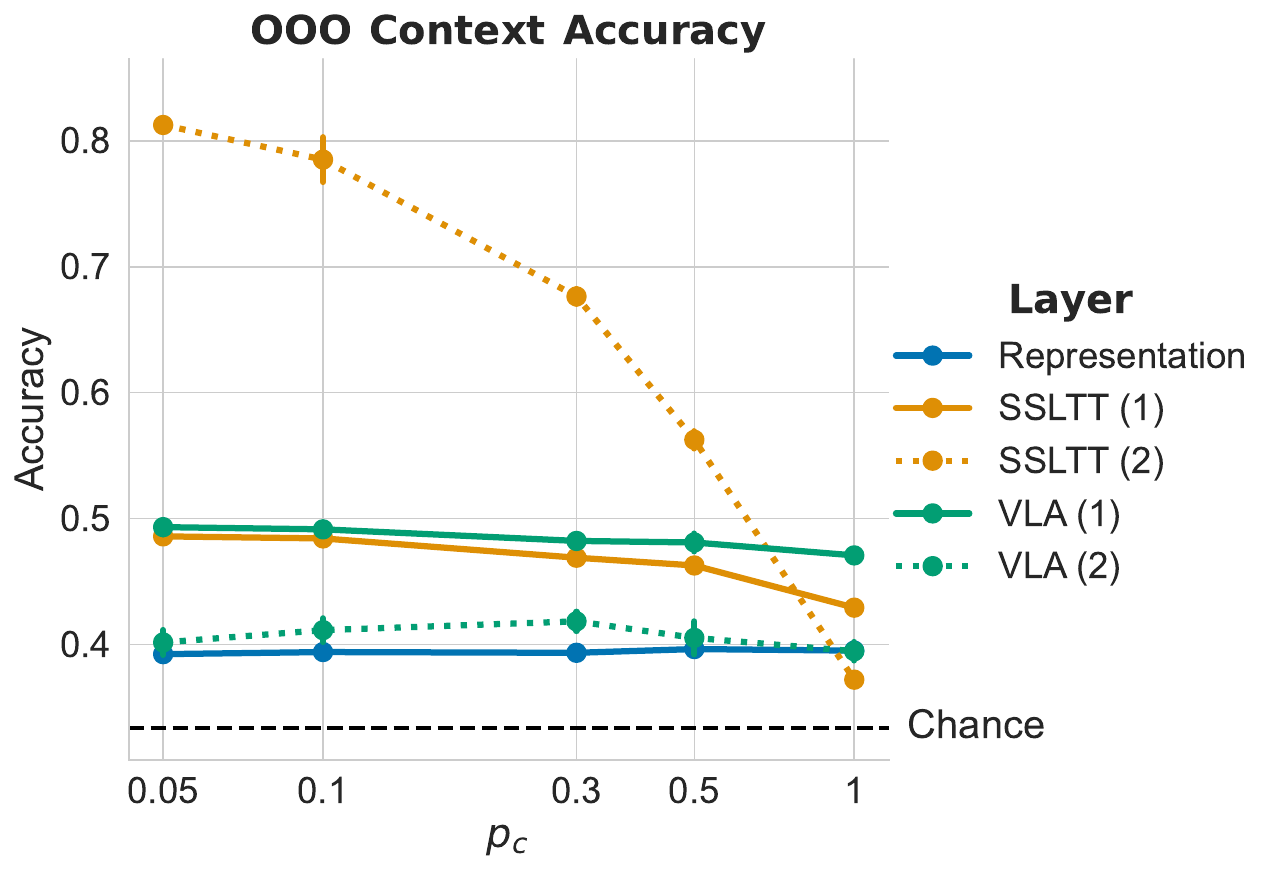}
     \caption{Odd-one-out test context accuracy against the probability of context transition $p_c$. Bars indicate standard deviation over two random seeds.}
    \label{fig:ooo_per_p}
\end{figure}

In this section, we examine the conditions driving the emergence of context-based similarity judgments. In \figref{fig:ooo_per_p}, we observe that the OOO-accuracy based on the SSLTT(2) layer drops substantially as $p_c$ increases. Interestingly, the output of the vision encoder and the label projection heads also capture minor contextual information, irrespective of the temporal structure of the visual experience driven by $p_c$.


\section{Conclusion}

We investigated how the temporal and multi-modal structure of first-person experience may shape object representations and instill semantic structure into them. We created a temporal sequence of real-world observations of objects and learnt visual representations using a bio-inspired learning model that simulates visuo-language alignment and alignment of successive experiences. Our results demonstrate that the model tends to cluster together objects co-occurring in the same, e.g., ``kitchen'', context. In contrast with the commonly optimal layer for category and instance recognition (middle layer), we find that the optimal layer for showing a context-wise structure is located in a higher neural network layer. Interestingly, this observation is consistent with recordings from human brains: category recognition is commonly associated with the ventral stream \cite{kruger2012deep} while the higher-level parahippocampal cortex encodes information about contexts \cite{aminoff2013role}. In our analysis, this correlates with an increased sparsity of the layer, in line with the hypothesis that the brain gradually compresses visual inputs \cite{barlow1961possible}. In addition, we found that the temporal slowness objective leverages spatio-temporal visual co-occurrences, whereas visuo-language alignment only focuses on spatial visual co-occurrences.

We made simplifying assumptions to study the conditions of emergence of context structures in the representational space. This came at the cost of realism. Our temporal sequences contain visually non-smooth transitions between objects and used backgrounds do not always correspond to a ``typical'' background, \textit{i.e.} objects were sometimes recorded in a shop. Furthermore, our learning model assumes that a caregiver always utters the word associated with the object being viewed. In the future, we plan to better approximate humans' egocentric experience with a continuous video stream enhanced with simulated gaze positions and realistic language input. We also plan to directly compare the representations to human representations, following previous studies \cite{turini2022hierarchical,bonner2021object}.



\section*{Acknowledgment}
 This work was funded by the Deutsche Forschungsgemeinschaft (DFG project 5368 ``Abstract REpresentations in Neural Architectures (ARENA)''), as well as the project ``The Adaptive Mind'' funded by the Excellence Program of the Hessian Ministry of Higher Education, Science, Research and Art (HMWK). We gratefully acknowledge support from GENCI–IDRIS (Grant 2022-AD011013678) and Goethe-University (NHR Center NHR@SW) for providing computing and data-processing resources needed for this work. Jochen Triesch was supported by the Johanna Quandt foundation.

\bibliography{references}
\bibliographystyle{IEEEtranS}
\end{document}